%% file: main.tex
\documentclass[12pt, letter]{article}
\input{Preamble/preamble}

\usepackage{subfig}
\usepackage{caption}
\usepackage{hyperref}

\usepackage{xspace}

\usepackage{graphicx}
\usepackage{algorithm}
\usepackage{caption}

\usepackage{booktabs}
\usepackage{bbm}
\usepackage{wrapfig}
\usepackage[noend]{algpseudocode}
\algnewcommand{\IfThenElse}[3]{
  \State \algorithmicif\ #1\ \algorithmicthen\ #2\ \algorithmicelse\ #3}
\algnewcommand{\IfThen}[2]{
  \State \algorithmicif\ #1\ \algorithmicthen\ #2}
\algrenewcommand\algorithmicrequire{\textbf{Input:}}
\algrenewcommand\algorithmicensure{\textbf{Output:}}

\begin{document}
\title{Accelerating Discovery in Natural Science Laboratories with AI and Robotics: Perspectives and Challenges from the 2024 IEEE ICRA Workshop, Yokohama, Japan}
\author{Andrew I. Cooper$^{\ddagger}$$^{1}$, Patrick Courtney$^{\ddagger}$$^{2}$, Kourosh Darvish$^{\dagger}$$^{3,4}$$^*$,\\
Moritz Eckhoff$^{\dagger}$$^{5}$, Hatem Fakhruldeen$^{\dagger}$$^{1}$, Andrea Gabrielli$^{\dagger}$$^{5}$, Animesh Garg$^{\ddagger}$$^{4}$,\\
Sami Haddadin$^{\ddagger}$$^{5}$, Kanako Harada$^{\ddagger}$$^{6}$, Jason Hein$^{\ddagger}$$^{7}$, Maria Hübner$^{\ddagger}$$^{8}$,\\
Dennis Knobbe$^{\dagger}$$^{5}$$^*$, Gabriella Pizzuto$^{\dagger}$$^{1}$$^*$, Florian Shkurti$^{\dagger}$$^{4}$, Ruja Shrestha$^{\ddagger}$$^{9}$,\\
Kerstin Thurow$^{\ddagger}$$^{10}$, Rafael Vescovi$^{\ddagger}$$^{11}$, Birgit Vogel-Heuser$^{\ddagger}$$^{12}$, Ádám Wolf$^{\ddagger}$$^{2,13,14}$,\\
Naruki Yoshikawa$^{\dagger}$$^{15}$, Yan Zeng$^{\ddagger}$$^{16}$, Zhengxue Zhou$^{\dagger}$$^{1}$, Henning Zwirnmann$^{\dagger}$$^{5}$\\
 \\
\normalsize{The authors are listed alphabetically.}\\
\normalsize{$^{\dagger}$ Workshop Organizers}\\
\normalsize{$^{\ddagger}$ Workshop Speakers}\\
\normalsize{$^{1}$ University of Liverpool, UK}\\
\normalsize{$^{2}$ SiLA Consortium, Switzerland}\\
\normalsize{$^{3}$ Acceleration Consortium, Canada}\\
\normalsize{$^{4}$ University of Toronto, Canada}\\
\normalsize{$^{5}$ Technical University of Munich, Germany; TUM School of Computation, Information, and Technology,}\\
\normalsize{Department of Computer Engineering, Chair of Robotics and Systems Intelligence;}\\
\normalsize{Munich Institute of Robotics and Machine Intelligence (MIRMI)}\\
\normalsize{$^{6}$ University of Tokyo, Japan}\\
\normalsize{$^{7}$ University of British Columbia, Canada}\\
\normalsize{$^{8}$ Roche Diagnostics GmbH, Germany}\\
\normalsize{$^{9}$ Unchained Labs, USA}\\
\normalsize{$^{10}$ Center for Life Science Automation, University of Rostock, Germany}\\
\normalsize{$^{11}$ Argonne National Laboratory, USA}\\
\normalsize{$^{12}$ Technical University of Munich, Germany; TUM School of Engineering and Design,}\\
\normalsize{Department of Mechanical Engineering, Institute of Automation and Information Systems;}\\
\normalsize{Munich Institute of Robotics and Machine Intelligence (MIRMI)}\\
\normalsize{$^{13}$ Takeda, Switzerland}\\
\normalsize{$^{14}$ Óbuda University, Hungary}\\
\normalsize{$^{15}$ Institute of Science Tokyo, Japan}\\
\normalsize{$^{16}$ Florida State University, USA}\\
\normalsize{$^{*}$ Corresponding authors; E-mail: Kourosh.Darvish@utoronto.ca,}\\ \normalsize{Dennis.Knobbe@tum.de, Gabriella.Pizzuto@liverpool.ac.uk}\\
}


\baselineskip24pt

\maketitle
\begin{sciabstract}
Science laboratory automation enables accelerated discovery in life sciences and materials. However, it requires interdisciplinary collaboration to address challenges such as robust and flexible autonomy, reproducibility, throughput, standardization, the role of human scientists, and ethics. This article highlights these issues, reflecting perspectives from leading experts in laboratory automation across different disciplines of the natural sciences.
\end{sciabstract}



\import{tex/}{Introduction}
\import{tex/}{Theme1}

\import{tex/}{Theme2}
\import{tex/}{Theme3}

\import{tex/}{Theme4}

\import{tex/}{Theme5}
\import{tex/}{Theme6}
\import{tex/}{Conclusions}

\bibliography{Bibliography/Bibliography}
\bibliographystyle{Science}

\import{tex/}{Acknowledgments}

\newpage
\import{tex/}{FiguresAndTables}

\FloatBarrier 
\setcounter{page}{1} 
\setcounter{figure}{0} 
\setcounter{equation}{0} 


\vfill

\newpage
\appendix
\setcounter{page}{1} 
\setcounter{figure}{0} 


\end{document}

%% file: Preamble/preamble.tex
\usepackage{amssymb}
\usepackage{amsmath} 
\usepackage{commath}
\usepackage{mathtools}
\usepackage{IEEEtrantools}

\usepackage{import}
\usepackage{paralist}
\usepackage{svg}
\usepackage{textcomp}
\usepackage{gensymb}
\usepackage{dsfont}
\usepackage{subfig}
\usepackage{graphics} 
\usepackage{epsfig} 
\usepackage[normalem]{ulem}
\usepackage[binary-units]{siunitx}
\usepackage{bm}
\usepackage[shortlabels]{enumitem}
\usepackage{upgreek}
\usepackage{todonotes}
\usepackage{scicite}
\usepackage{times}
\usepackage{xurl}
\usepackage{multirow}
\usepackage{makecell}
\usepackage{tabularx}
\usepackage{booktabs}
\usepackage[percent]{overpic}
\usepackage{xcolor}
\usepackage{transparent}
\usepackage{tikzscale}
\usepackage[htt]{hyphenat}
\usepackage{nameref}
\usepackage{chngcntr}
\usepackage{hyperref}
\usepackage{placeins}
\usepackage{spverbatim}
\usepackage{listings}

\definecolor{codegreen}{rgb}{0,0.6,0}
\definecolor{codegray}{rgb}{0.4,0.4,0.4}
\definecolor{codepurple}{rgb}{0.5,0,0.9}
\definecolor{backcolour}{rgb}{0.95,0.95,0.95}

\lstdefinestyle{mystyle}{
    backgroundcolor=\color{backcolour},   
    commentstyle=\color{codegreen},
    keywordstyle=\color{magenta},
    numberstyle=\tiny\color{codegray},
    stringstyle=\color{codepurple},
    basicstyle=\fontsize{10}{10}\selectfont\ttfamily\ttfamily,
    breakatwhitespace=false,         
    breaklines=true,
    breakindent=0pt,
    captionpos=b,                    
    keepspaces=true,                 
    numbers=none,                    
    numbersep=5pt,                  
    showspaces=false,                
    showstringspaces=false,
    showtabs=false,                  
    tabsize=3
}
\newcommand\blfootnote[1]{%
    \bgroup
    \renewcommand\thefootnote{\fnsymbol{footnote}}%
    \renewcommand\thempfootnote{\fnsymbol{mpfootnote}}%
    \footnotetext[0]{#1}%
    \egroup
}

\newenvironment{sciabstract}{%
\begin{quote} \bf}
{\end{quote}}

\date{}

%% file: tex/Introduction.tex
\section*{Introduction}
\label{sec:introduction}



Fundamental breakthroughs across many scientific disciplines are becoming increasingly rare~\cite{park2023papers}. At the same time, challenges related to the reproducibility and scalability of experiments, especially in the natural sciences~\cite{Baker2016,musslick2024automating}, remain significant obstacles. For years, automating scientific experiments has been viewed as the key to solving this problem. However, existing solutions are often rigid and complex, designed to address specific experimental tasks with little adaptability to protocol changes. With advancements in robotics and artificial intelligence, new possibilities are emerging to tackle this challenge in a more flexible and human-centric manner. 
However, it remains unclear how to apply these innovations effectively to fundamentally accelerate knowledge generation and discovery in semi-structured environments across the natural sciences, spanning the life and physical sciences.

This article presents and discusses the results of the ``Accelerating Discovery in Natural Science Laboratories with AI and Robotics" workshop~\cite{Workshop2024ScienceLab}, held during the 2024 IEEE International Conference on Robotics and Automation (ICRA) in Yokohama, Japan. 
The workshop facilitated knowledge exchange between academic researchers exploring robotics for science automation and industry partners in sectors such as robotics, pharmaceutical and lab automation who could benefit from these innovations.
It featured experts in robotics alongside professionals from other academic fields and various industries, including materials science (discovery and synthesis), life sciences, and process systems engineering. An overview of some pioneering works from the workshop participants is illustrated in Fig.~\ref{fig:example_works}.
This event fostered unique synergies between academia and industry, facilitating collaborations to overcome challenges and drive discoveries in the natural sciences.

Recently, several articles on lab automation have been published, including two surveys on self-driving labs in chemistry and materials science ~\cite{abolhasani2023rise, tom2024self} and a viewpoint on science lab automation~\cite{angelopoulos2024transforming}. Although these works introduce the current status, components, and categorise lab automation by metrics such as autonomy, generalisation, hardware, and software, our viewpoint primarily highlights the current challenges in science lab automation based on insights from world-leading experts who delivered exciting talks on their recent works at the workshop~\cite{Workshop2024ScienceLab}.
Speakers at the workshop represented a diverse range of institutions, companies, genders, ethnicities and career stages, demonstrating our commitment to providing a viewpoint with equity and representation across the field.

The remainder of this article is visualised in Fig.~\ref{fig:themes}. It comprises six themes, starting with Theme I which draws the picture of the quest and the challenges of automating natural sciences. Theme II contains ideas and opinions on the role of digital twins and simulators in accelerating scientific discovery. As a long-standing topic of debate, Theme III explores the question of the level of autonomy required for flexible experimental setups. Theme IV focuses on foundation models and generative AI for general-purpose robots in the context of automated scientific discovery. Theme V addresses standardization and its application in the field. Finally, theme VI examines reproducibility, safety, risk, sustainability, and ethical considerations in laboratory automation and knowledge discovery. The concluding section of this viewpoint article summarises the key points discussed.

%% file: tex/Theme1.tex
\section*{Theme I: The Quest and Challenges in Natural Science Automation}
\label{sec:themeI}






The quest for science automation seeks to transform how we approach discovery and experimentation, enabling researchers to tackle complex problems with greater efficiency, throughput, precision, and reproducibility, ultimately accelerating the pace of experiments. The goal is to shift from manual, labour-intensive procedures to intelligent, autonomous systems capable of iterating experiments and gathering high-quality, standardised data, eliminating uncertainty caused by human variability. Autonomous labs should combine insights from existing scientific knowledge with large datasets, these systems can make guided, real-time decisions that may even outperform human experts. Additionally, such autonomous systems can improve security, safety, and traceability for compliance. To build such frameworks, interoperability and integration of legacy systems with new technologies, or among solutions from different vendors, is essential.
Furthermore, advancements in robotics are needed to enable flexible solutions through a combination of precise and generalizable perception and manipulation.
With technologies that operate without demanding constant attention, scientists can focus on problem-solving and research, while systems handle complex tasks such as onboarding new experimental ``recipes'' defined by human experts, with minimal need for reprogramming or customization by the end user. This interaction across multiple dimensions fosters collaboration and knowledge sharing within and across scientific communities.

A key challenge in science automation is integrating heterogeneous hardware and software tools into a unified system, as labs rely on various instruments designed for different tasks, making a one-size-fits-all solution impractical. Building and maintaining such an integrated, interoperable system requires a cohesive framework for data collection, control, and decision-making. This effort demands multidisciplinary teams, significant capital, time investment, and automation-friendly tools. To tackle these challenges, standardisation of communication protocols, interfaces, and open-source system designs across industries and research communities is essential for enabling the development of autonomous labs. This approach encourages a culture of sharing knowledge and expertise, facilitating innovation and adaptation across various research settings and product designs through partnerships and shared infrastructure.

The effectiveness of automation in scientific experimentation hinges on its flexibility, robustness, and scalability.
A key challenge, however, is to optimize experiments for time and resource efficiency while ensuring that scaling up does not create exclusivity. To democratize lab automation and prevent it from becoming an ``exclusive club,'' we must design systems that promote broad participation rather than depending solely on massive, centralized infrastructures like the European Organisation for Nuclear Research (CERN).
As automation systems evolve, a critical question arises: should the automation adapt to the processes, or should the processes conform to the available automation options? This question underscores the need for flexibility and reconfigurability, especially as experimental setups change frequently, necessitating adaptable systems that can manage evolving workflows and configurations while adhering to safety constraints. 
Automation systems must be versatile enough to adjust protocols on the fly, particularly in dynamic conditions encountered during early-stage discovery or process optimization. Furthermore, addressing emerging scientific challenges involves developing robust automation solutions that can support large-scale, long-term experiments while integrating with existing hardware. Reliability can be achieved through error detection that allows systems to autonomously respond to issues and avoid disruptions in experiments. 
Moreover, a modular design for lab automation systems can facilitate integration and reconfiguration, thereby reducing setup complexity and increasing flexibility. Handling such a complex and heterogeneous system requires a universal workflow management orchestrator that is both versatile and user-friendly, enabling laboratories to implement effective automation without extensive technical knowledge.

Given the exponential growth of data generated by advanced analytical techniques and a diverse set of sensors in automated labs, efficient and secure data management is essential. The challenge lies in ensuring data integrity and translating complex data streams into actionable insights for autonomous decision-making, especially when dealing with subtle changes, such as reaction dynamics within chemical processes. As automation scales, the ability to convert raw data into real-time insights becomes crucial; without robust analysis frameworks, even sophisticated automated robotics labs may struggle to provide value. Leveraging machine learning and artificial intelligence to manage and interpret the expanding data landscape, as well as to evaluate performance through multi-modal proxies, can facilitate faster insights and predictive decision-making. Successfully addressing this challenge will unlock the potential of autonomous platforms, driving more effective experimentation and accelerating scientific breakthroughs.

%% file: tex/Theme2.tex
\section*{Theme II: The Role of Digital Twins and Simulators for Accelerating Scientific Discovery}
\label{sec:themeII}



Innovative digital technologies from industrial manufacturing, such as digital twin concepts, provide promising solutions to many manufacturing challenges. But can they also be applied in scientific laboratories to accelerate discoveries? 

In a factory setting, digital twins are revolutionizing variant management in product line engineering by utilizing modern data handling methods to enhance traditional product lifecycle management. By definition, they encompass both engineering models and knowledge, including simulations, documents, models, and automation control code, as well as operational data such as process or alarm data collected during operation \cite{VogelHeuser2024Methods}. In essence, the digital twin comprises every digitally available artifact of the system. 

When this technology is applied to a scientific environment, it can be said that as the experiments evolve, so do their digital twins. Similar to the factory setting, in the best case they capture all relevant information, facilitate the reproduction of experiments, and at the same time provide valuable insights for new scientific discoveries. They enable a meaningful observation of the available system data and state, creating a feedback loop that helps to control and optimize the process and the system itself. Furthermore, digital twins can reveal from gathered data relationships between experimental parameters and results, e.g. in the form of cause-and-effect graphs.

Scientists can also use them as a foundation to identify and address critical conditions and errors. By utilizing simulation and AI as part of a digital twin, they can model, analyze, and optimize experiments, proposing effective solutions before applying them in real-world scenarios. Overall, this enhances decision-making and the explainability of complex systems, highlighting the value of digital twins. 

From an implementation perspective, the greatest challenge is creating interfaces and asset representations that seamlessly integrate devices and containers into a digital twin system. In this context, standards like Standardization in Lab Automation (SiLA2) \cite{bar2012sila,hinkel2023tecan} or Open Platform Communication – Unified Architecture (OPC UA) \cite{Leitner2006opcua} are essential for enabling the technology to function effectively at scale. Additionally, the use of appropriate information models, such as ontologies, is vital for accurately representing the interdependencies among assets within the digital ecosystem.

In summary, even though the deployment of digital twins comes with challenges that require careful management and strategic planning, they bring substantial benefits, including cost reduction, improved accuracy of scientific results, and personalized solutions.

%% file: tex/Theme3.tex
\section*{Theme III: Level of Autonomy Towards Flexible Experimental Setups}
\label{sec:themeIII}


The degree of autonomy in science automation and knowledge discovery is a topic of debate.
From an academic perspective, the goal is not to replace scientist’s contributions, but to complement and enhance them by combining human creativity with machine efficiency. This creates a symbiotic feedback loop, with scientists and machines working together to accelerate knowledge discovery. Machines evolve differently; rather than replicating all human capabilities, the aim is to develop machines that help conduct experiments and push beyond human limitations.
This is where trained scientists play a crucial role in interpreting and synthesizing meaningful results from automated data. There is a great need for workflows that effectively utilize this knowledge and are led by experts who can fully understand and apply the results of the experiments. However, in some cases, autonomous systems themselves can serve as a source of ``inspiration'' for scientists, potentially leading to innovative discoveries that humans might not have achieved on their own.

One of the most challenging factors from the academic view is the robustness of the automation system. It is a decisive factor that can, but does not have to, compromise the system's flexibility. In general, the failure rates for individual modules in the workflow need to be very low. For instance, in an autonomous workflow involving 20 operations, a 99\% success rate per operation leads to an overall success rate of only 82\%, meaning the workflow will fail approximately once every five runs. Ultimately, robustness can only be achieved through a modular approach, with thorough testing of each individual module. Enhancing the reliability of automated systems still depends on human involvement in the experimental workflow through a human-in-the-loop approach. 

The industrial perspective supports this view, emphasizing that technology should serve as an assistive tool offering high flexibility in both throughput and handling. It should empower process experts to share and expand their knowledge of the experiment. Interfaces should be application-based, removing the need to directly program the robot or automated system. Instead, routines can be created using drag-and-drop applications on a laptop or tablet. As technologies advance, the need for human involvement may decrease; however, business requirements and the maturity of the technology may still necessitate human participation in certain processes. In some cases, humans may act as input providers or be responsible for verifying outputs.

Although a human-in-the-loop approach may not fully achieve the goals or benefits of fully automated labs — such as generating high-quality data in large quantities without human uncertainty — it does provide advantages, including reduced startup effort, complexity, and overall risks. While fully autonomous laboratories may be a future goal, current automation systems mainly function as assistive tools. A key concern of fully autonomous laboratories is the need for flexibility, particularly the ability to adjust experimental parameters on a fundamental level during the experiment as new insights emerge. 

In summary, by establishing a complementary relationship between humans and machines, the integration and advancement of AI and robotics in natural science laboratories will not be something to be afraid of. Instead, it should be seen as a means to collaboratively expand the frontiers of science. For highly automated labs, human ingenuity is critical in designing, implementing, and supervising autonomous workflows, and AI will further increase comfort levels. Humans should maintain scientific inquiry throughout the workflow and strive to understand the results, independently verify the outcomes, and provide timely reports to the broader scientific community.

%% file: tex/Theme4.tex
\section*{Theme IV: Foundation Models and Generative AI for General-purpose Robots}
\label{sec:themeIV}

Generative artificial intelligence, especially foundation models like Large Language Models (LLMs) and Vision-Language Models (VLMs), plays a crucial role in transforming laboratory automation. These models are essential for enhancing data processing, hypothesis generation, experimental planning, and decision-making. LLMs are particularly effective for automating literature reviews, protocol generation, and documentation, as they can analyze and interpret large volumes of textual data. VLMs are great for image and data analysis, especially when it comes to handling multi-modal data, such as combining visual information with text. They can also enable users to engage with complex systems without needing in-depth technical knowledge by allowing researchers to interact with lab systems using natural language.

However, classical machine learning methods are still valuable for predictive modeling and identifying trends from historical data, while reinforcement learning (RL) shines in optimizing experimental processes and managing real-time decision-making in automated labs. RL's ability to dynamically adjust workflows makes it an excellent choice for applications that require flexibility and adaptability.

The choice of the optimal method depends on the task requirements. Generalized models are versatile and can handle diverse tasks, making them ideal for environments with variable requirements. However, they often require significant computational resources. Specialized models, on the other hand, offer precision and efficiency for specific tasks and are more resource-efficient, making them suitable for labs with limited computational capacity. It is also worth noting that while foundation and generative models serve as non-explicit memories, our goal is to make knowledge as compact, quick, and accurate as possible while ensuring it is generalizable. As machine learning evolves, we will discover smarter and clearer ways to store knowledge and these models will continue to be pivotal in advancing laboratory automation.

%% file: tex/Theme5.tex
\section*{Theme V: Standardisation}
\label{sec:themeV}

Standardisation is key to achieve seamless system integration, support modularity, and streamlined setups. It would result in minimising the time spent on interfacing instruments from different vendors and building automated workflows. 
Consequently, this will lower entry costs and adoption barriers for many labs that have limited technical expertise in the areas of automation and system integration.
While the initial efforts for standardisation across different laboratories are tedious, the long-term benefits of such an endeavor will undoubtedly result in wider adoption of robotics, automation, and knowledge sharing. 
It is therefore crucial to emphasize that standardisation will promote and accelerate innovation through greater interoperability and reduced efforts when integrating and building systems. 
This will only be realisable through active collaboration and knowledge sharing of best practices among the key industrial players in the domain, especially knowledge transfer between the longer established companies with legacy products and proprietary interfaces and protocols, to small and medium size enterprises (SMEs) open to clean sheet design, that would benefit from widely available and tested standards and open frameworks. 
It is therefore imperative that active research labs and stakeholders, e.g., large pharmaceutical companies, play a leading role in promoting standardised communication protocols in their user-requirements specifications.
One way to achieve this is by connecting standards and best practices with tangible business values such as stronger cybersecurity protection, greater quality assurance, and facilitation of data handling and management by adopting the Findable, Accessible, Interoperable, and Reusable (FAIR) data principles~\cite{Wilkinson2016TheFG}. 

However, there are several bottlenecks that hinder the adoption of new standards.
In general, standardising requirements tends to be reactive, stemming from the slow-moving nature of the scientific community to adopt new standards on `already-working' technologies.
This can only be affected and put into practice through a top-down management approach and a digitally integrated scientific workforce. 
Furthermore, some instrument manufacturers build their business model around their unique, proprietary, non-standardised solutions; this makes it even harder for them to align with a universal standard where the benefit might not be obvious from day one. 
The field is also very domain-focused, where the end-users might have a widely different set of experiments that change over months and years. 
Hence, it becomes difficult to standardise across such a broad field, e.g. life sciences might not have the same user requirements as the materials community. 

In summary, the key benefits of standardisation are ease of implementation (potentially plug-and-play functionality) and interoperability; further adoption will also promote knowledge sharing and easier data transfer between scientists. 
Furthermore, taking a user-centric approach towards standardisation with all stakeholders playing an important role will ensure that user needs are met and increase overall adoption and usage.
Ultimately, it will come down to a combination of user demand, industry active collaboration, and ongoing development of standards (e.g., SiLA, OPC UA) to enhance interoperability and efficiency in lab automation.

%% file: tex/Theme6.tex
\section*{Theme VI: Reproducibility, Safety, Risk, Sustainability, Ethical Factors}
\label{sec:themeVI}

The inherent variability of manual experimentation in the natural sciences has led to concerns about the reproducibility of research findings. 
It is the consensus that laboratory automation has the potential to eliminate errors that are inherently common during manual execution, whilst simultaneously providing comprehensive audit trails and a wider range of stored data. 
Nonetheless, this could be at the expense of losing out on natural variation in experimental results that has historically led to breakthrough scientific discoveries. 

One of the most important factors that underpin the transformative change that robotics and automation can play in this field is unlocking the potential of carrying out dangerous and understudied reactions involving hazardous materials. 
Human safety could also be safeguarded through additional automation features, e.g., interlocks and shut-offs that halt operation when unsafe conditions are sensed. 
This can only be achieved through regulation and proper certification of new equipment; here, standardisation can play a key role, as having best practices already available would make it easier to put into action for improving overall safety. 
Without the correct regulations and standardisation methods in place, the inherent risk of deploying these systems would naturally increase. 

Automated systems are also placed towards increasing sustainability as in comparison to manual workflows. These systems are optimised towards improved efficiency and minimised usage of materials and energy. 
Furthermore, automated workflows could be run at much smaller scale due to the more rich information that can be collected, leading to potentially reduction in waste generation. 
However, this is in practice all speculative, and we argue that benchmarking efforts similar to the green lab initiative~\cite{GreenLabInitiative} with the correct metrics such as energy and material consumption would inform us whether this is correct. 
It is therefore timely and crucial to put into place practices towards tracking error rates and measuring variability and consistency between manual and automated experiments.

With an increased usage of this technology, it becomes evident that its end-use could lead to dual-usage that goes beyond the planned outcome. 
This could raise ethical concerns regarding the potential misuse of laboratory autonomous systems, and as such regulation and best practices are key towards mitigating such cases. 
Given the rapid pace of discovery stemming from novel automated methods, responsible research practices need to be continuously updated to meet the latest technology. 
Another vital ethical factor is to ensure the safety and health of laboratory-based scientists, who should be upskilled in the face of new automated systems.
This will ensure there is an increased uptake and sustainable development of laboratory automation platforms that is aligned to their needs and skills. 

By carefully considering and addressing these safety, sustainability, and ethical factors, we can ensure that lab automation is implemented responsibly and effectively to advance scientific progress while minimising risks and maximising benefits.

%% file: tex/Conclusions.tex
\section*{Conclusions}
\label{sec:conclusions}

Science lab automation has the potential to transform research across disciplines by making experimentation more precise, efficient, and reproducible. Achieving this vision requires overcoming challenges in integrating diverse instruments and managing vast data through AI-driven insights. Digital twins and foundation models offer promising tools to support scientists, enabling optimized experimental planning and data analysis. However, a successful future for autonomous labs must balance robust, modular design with flexibility, foster industry collaboration for standardization, and carefully address ethical concerns. With scientists in the loop, lab automation can become a powerful partner in accelerating science discovery and enabling fundamental breakthroughs.

%% file: tex/Acknowledgments.tex
\newpage
\section*{Acknowledgments}
We thank Al\'{a}n Aspuru-Guzik, Dejanira Araiza-Illan, Liwei Qi for the insightful discussions.

\paragraph{Authors contributions}
The workshop organizers drafted questions for the speakers based on the ICRA workshop panel discussion. The original draft of the paper was written by the organizers, incorporating responses from the speakers. All authors reviewed and proofread the manuscript.

\paragraph{Declaration of interests}
The authors declare no competing interests.

\paragraph{Funding}
This work was supported by the University of Toronto's Acceleration Consortium from the Canada First Research Excellence Fund, grant number CFREF-2022-00042, the Leverhulme Trust through the Leverhulme Research Centre for Functional Materials Design, and the Royal Academy of Engineering under the Research Fellowship Scheme. We gratefully acknowledge the funding of this work by the Alfried Krupp von Bohlen and Halbach Foundation. We gratefully acknowledge the funding of this work by the Deutsche Forschungsgemeinschaft through the Gottfried Wilhelm Leibniz Programme (award to Sami Haddadin; grant no. HA7372/3-1). We gratefully acknowledge the funding of robo.innovate by the Bavarian State Ministry for Economic Affairs, Regional Development and Energy (StMWi) and like to thank the initiative “Gründerland Bayern” for the continuous support. Funded by the Federal Ministry of Education and Research (BMBF) and the Free State of Bavaria under the Excellence Strategy of the Federal Government and the Länder. This work was supported in part by Laboratory Directed Research and Development funds at Argonne National Laboratory from the U.S. Department of Energy under Contract DE-AC02-06CH11357.

%% file: tex/FiguresAndTables.tex
\section*{Figures and Tables}


\begin{center}
    \centering
    \maketitle
    \includegraphics[width=1.0\textwidth]{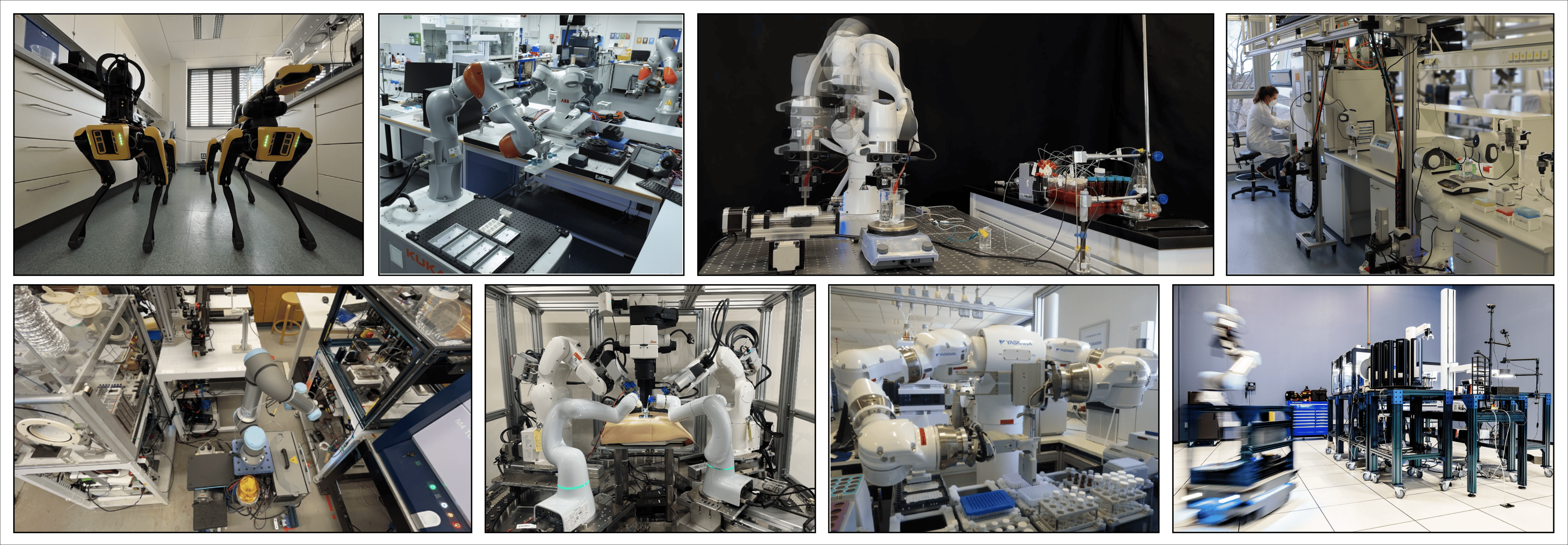}
    \captionof{figure}{\textbf{ Examples of science lab automation; from top left to bottom
right corner:~\cite{wolf, amy, darvish2024organaroboticassistantautomated, dennis, hein, kanako, rostock, vescovi2023towards}  .}}
    \label{fig:example_works}
\end{center}


\begin{center}
    \centering
    \maketitle
    \includegraphics[width=1.0\textwidth]{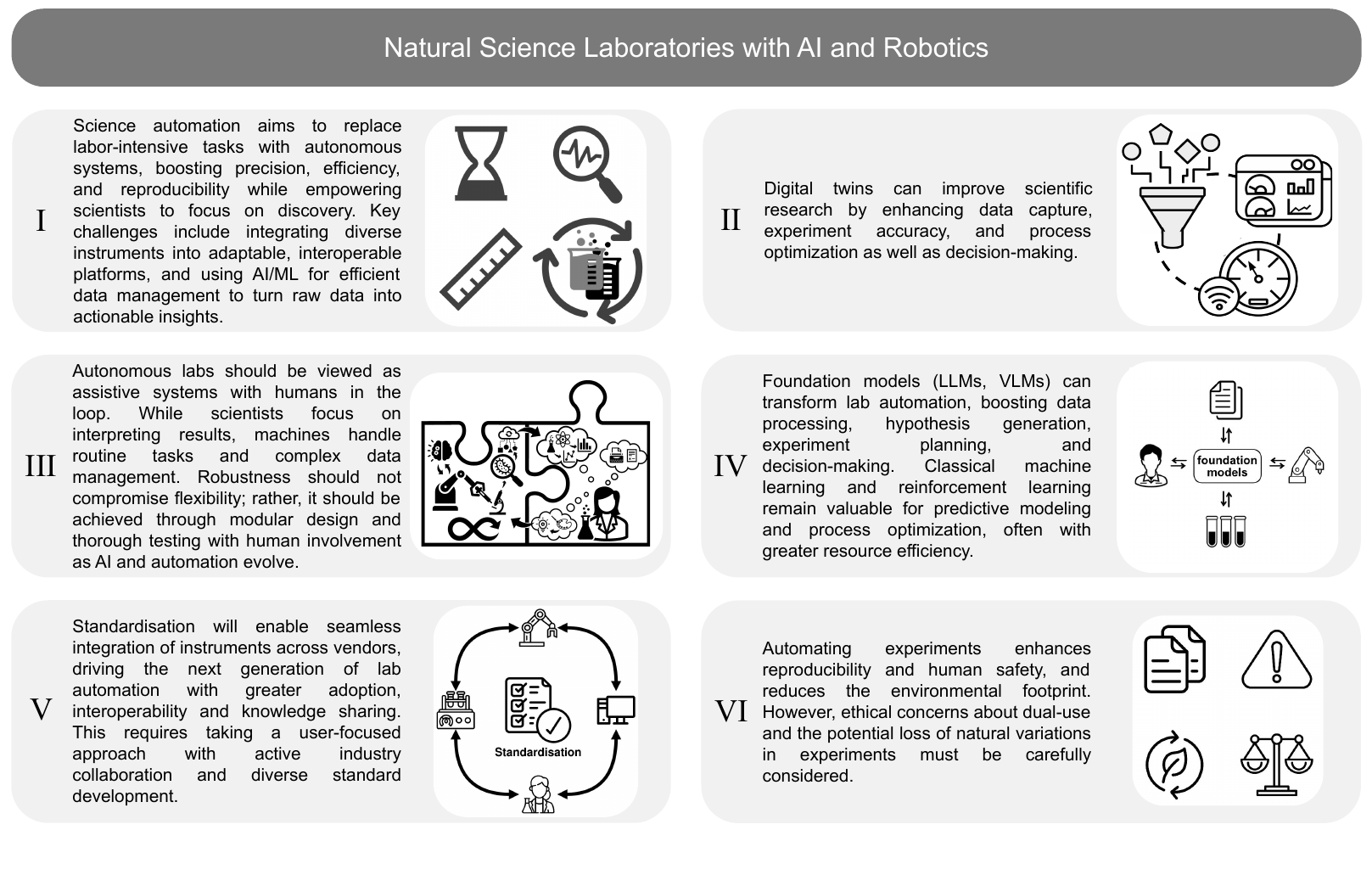}
    \captionof{figure}{\textbf{Overview of perspectives and challenges in science lab automation using robotics and AI.}}
    \label{fig:themes}
\end{center}

\newpage